\documentclass[conference]{IEEEtran}
\usepackage[table]{xcolor}
\usepackage{eda} 
\usepackage{microtype}
\usepackage{graphicx}
\usepackage{booktabs} 
\usepackage{amssymb}
\usepackage{amsmath}
\usepackage{amsthm}
\usepackage{enumitem}
\usepackage{physics}
\usepackage{rotating}
\usepackage{mathtools}
\usepackage{algorithm}
\usepackage[noend]{algpseudocode}
\usepackage{caption}
\usepackage{subcaption}
\usepackage[nocompress]{cite}
\usepackage{balance}
\usetikzlibrary{positioning,calc}

\DeclareRobustCommand*{\IEEEauthorrefmark}[1]{\raisebox{0pt}[0pt][0pt]{\textsuperscript{\footnotesize\ensuremath{\ifcase#1\or\bullet\or\circ\or*\or \dagger\or \ddagger\or%
				\mathsection\or \mathparagraph\or \|\or **\or \dagger\dagger%
				\or \ddagger\ddagger \else\textsuperscript{\expandafter\romannumeral#1}\fi}}}}

\usepackage{tikz}
\usetikzlibrary{snakes,arrows,shapes,positioning,fit,calc,patterns}
\usepackage{pgfplots}
\tikzset{cross/.style={cross out, draw=black, fill=none, minimum size=2*(#1-\pgflinewidth), inner sep=0pt, outer sep=0pt}, cross/.default={2pt}}
\usetikzlibrary{external}
\usepackage{nicefrac}

\newtheorem{theorem}{Theorem}

\newtheorem{lemma}{Lemma}
\newtheorem{proposition}{Proposition}

\theoremstyle{definition}
\newtheorem{definition}{Definition}

\DeclareMathOperator*{\argmax}{arg\,max}

\newcommand{\defemph}[1]{\textbf{#1}}
\newcommand{\mapping}[3]{#1 \! : #2 \to #3}
\newcommand{\card}[1]{| #1 |}
\newcommand{\with}{\! : \,}
\newcommand{\given}{\, | \,}

\newcommand{\domain}[1]{\ensuremath{V_{#1}}}
\newcommand{\domains}[1]{\ensuremath{S_{#1}}}
\newcommand{\expected}[1]{\ensuremath{\mathbb{E}\big[#1\big]}}

\newcommand{\cI}{\mathcal{I}}
\newcommand{\cP}{\mathcal{P}}
\newcommand{\cR}{\mathcal{R}}
\newcommand{\cS}{\mathcal{S}}

\newcommand{\cX}{\mathcal{X}}

\newcommand{\cZ}{\mathcal{Z}}

\newcommand{\D}{\mathbf{D}}

\newcommand{\hatp}{\ensuremath{\hat{p}}}

\newcommand{\hH}{\ensuremath{\hat{H}}}
\newcommand{\mut}{\ensuremath{I}}
\newcommand{\hmut}{\ensuremath{\hat{I}}}
\newcommand{\hmuto}{\ensuremath{\hat{I}_0}}

\newcommand{\mo}{\ensuremath{m_0}}

\newcommand{\Eo}[1]{E_0 [#1]}

\newcommand{\tc}{\ensuremath{W}}
\newcommand{\htc}{\ensuremath{\hat{W}}}

\newcommand{\ntc}{\ensuremath{w}}
\newcommand{\hntc}{\ensuremath{\hat{w}}}
\newcommand{\hntco}{\ensuremath{\hat{w}_{0}}}
\newcommand{\hntcbo}{\ensuremath{\hat{w}_{\bar{0}}}}
\newcommand{\hntcbd}{\ensuremath{\hat{w}_{\bar{\bar{0}}}}}

\newcommand{\mbo}{\ensuremath{m_{\bar{0}}}}
\newcommand{\mbop}{\ensuremath{m_{\bar{\bar{0}}}}}

\newcommand{\btc}{\ensuremath{\bar{w}}}
\newcommand{\tcbound}{\ensuremath{\bar{W}}}

\newcommand{\bhntcd}{\ensuremath{\bar{w}_{\bar{\bar{0}}}}}
\newcommand{\btcmon}{\ensuremath{\bar{w}_{\bar{\bar{0}}\text{mon}}}}
\newcommand{\btcref}{\ensuremath{\bar{w}_{\bar{\bar{0}}\text{ref}}}}

\newcommand{\tzero}{\ensuremath{t_0}}
\newcommand{\tbzero}{\ensuremath{t_{\bar{0}}}}
\newcommand{\tbzerop}{\ensuremath{t_{\bar{\bar{0}}}}}

\newcommand{\bnb}{\textsc{BnB}\xspace}

\newcommand{\BEAM}{\textsc{Greedy}}
\newcommand{\Beam}{\textsc{Greedy}\xspace}
\newcommand{\branch}{\mathbf{r}}
\newcommand{\oest}{\bar{f}}

\newcommand{\lmer}{\subseteq_{\scriptsize H}}
\graphicspath{ {./} }

\newcommand{\giturl}{\url{https://github.com/pmandros/wodiscovery}}

\def\BibTeX{{\rm B\kern-.05em{\sc i\kern-.025em b}\kern-.08em
		T\kern-.1667em\lower.7ex\hbox{E}\kern-.125emX}}
\begin{document}
	
	\title{Discovering Reliable Correlations\\in Categorical Data}
	
	\author{%
		\IEEEauthorblockN{Panagiotis Mandros\IEEEauthorrefmark{1}, Mario Boley\IEEEauthorrefmark{2}, Jilles Vreeken\IEEEauthorrefmark{3}}
		\IEEEauthorblockA{\IEEEauthorrefmark{1}Max Planck Institute for Informatics, Saarbr\"ucken, Germany\\
			\IEEEauthorrefmark{2}Monash University, Melbourne, Australia\\
				\IEEEauthorrefmark{3}CISPA Helmholtz Center for Information Security,
				Saarbr\"ucken, Germany\\
			pmandros@mpi-inf.mpg.de, mario.boley@monash.edu, jv@cispa.saarland} 
	}

	\maketitle
	
	\begin{abstract}
		
In many scientific tasks we are interested in discovering whether there exist any correlations in our data. This raises many questions, such as how to \emph{reliably} and \emph{interpretably} measure correlation between a multivariate set of attributes, how to do so without having to make assumptions on distribution of the data or the type of correlation, and, how to \emph{efficiently} discover the top-most reliably correlated  attribute sets from data. In this paper we answer these questions for discovery tasks in categorical data. 

In particular, we propose a corrected-for-chance, consistent, and efficient estimator for normalized total correlation, by which we obtain a reliable, naturally interpretable, non-parametric measure for correlation over multivariate sets. For the discovery of the top-$k$ correlated sets, we derive an effective algorithmic framework based on a tight bounding function. This framework offers exact, approximate, and heuristic search. Empirical evaluation shows that already for small sample sizes the estimator leads to low-regret optimization outcomes, while the algorithms are shown to be highly effective for both large and high-dimensional data. Through two case studies we confirm that our discovery framework identifies interesting and meaningful correlations.

	\end{abstract}
	
	
	\section{Introduction}
\label{sec:intro}

Most data are multi-dimensional, and identifying lower-dimensional correlated subsets of features is a fundamental aspect in many data analysis tasks. Such correlations are useful in many application, including  the discovery of treatments for diseases, network intrusions, earthquakes etc.~\cite{ke:2008:cor}. It is important that we can measure correlations over \emph{multivariate} sets of features, as genes for example may reveal only a weak correlation with a disease when considered individually, while the correlation over a group of genes can be very strong~\cite{zhang2008mining}. It is also important that our measure is
\emph{reliable}, such that we do not discover spurious correlations, that it is \emph{interpretable}, such that we know what a value means, and \emph{non-parametric}, such that we do not need to assume anything about the data distribution or type of correlation. Last, but not least, as we need to be able to efficiently discover the top-$k$ most correlated sets from possibly large quantities of data, we require an effective~search~framework~for~it. 

Information theory, with the tools to quantify uncertainty, offers an attractive framework to do exactly this. We build on the concept of \defemph{total correlation}, the multivariate extension of mutual information, which non-parametrically quantifies the amount of shared information in a set of random variables~\cite{watanabe:1960:tc}. 
Without appropriate normalization, however, scores over sets of different cardinalities are not comparable, which is a problem when searching for the top-most correlations~\cite{nguyen2016universal,wang2017unbiased}. We hence consider \textbf{normalized total correlation}, which does not only address this, but is also interpretable: a score of $0$ means the random variables in a set are statistically independent, and a score of $1$ that  there exists a variable that ``explains" all~others. 

\begin{figure}[t]
	\centering
	\begin{minipage}[b]{0.32\linewidth}
		\centering
		\centering
		\includegraphics[width=0.65\linewidth]{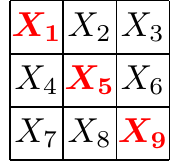}
		\vspace{-0.15cm}
		\caption*{top-$1$}
	\end{minipage}
	\begin{minipage}[b]{0.32\linewidth}
		\centering
			\includegraphics[width=0.65\linewidth]{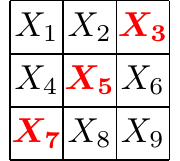}
		\vspace{-0.15cm}
		\caption*{top-$2$}
	\end{minipage}
	\begin{minipage}[b]{0.32\linewidth}
		\centering
				\includegraphics[width=0.65\linewidth]{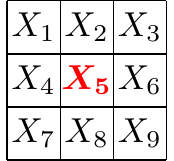}
		\vspace{-0.15cm}
		\caption*{top-$3$}
	\end{minipage}\\ \vspace{0.1cm}
	\begin{minipage}[b]{0.32\linewidth}
		\centering
			\includegraphics[width=0.65\linewidth]{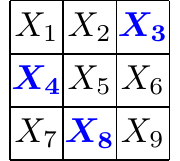}
		\vspace{-0.15cm}
		\caption*{top-$4$}
	\end{minipage}
	\begin{minipage}[b]{0.32\linewidth}
		\centering
				\includegraphics[width=0.65\linewidth]{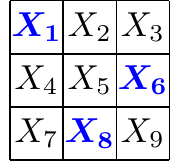}
		\vspace{-0.15cm}
		\caption*{top-$5$}
	\end{minipage}
	\begin{minipage}[b]{0.32\linewidth}
		\centering
				\includegraphics[width=0.65\linewidth]{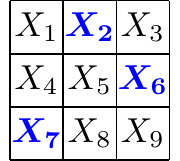}
	\vspace{-0.15cm}
	\caption*{top-$6$}
\end{minipage} \\ \vspace{0.2cm}
\begin{minipage}[b]{0.32\linewidth}
	\centering
			\includegraphics[width=0.65\linewidth]{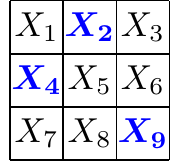}
\vspace{-0.15cm}
\caption*{top-$7$}
\end{minipage}
\begin{minipage}[b]{0.32\linewidth}
	\centering
		\includegraphics[width=0.65\linewidth]{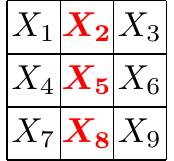}
	\vspace{-0.15cm}
	\caption*{top-$8$}
\end{minipage}
\begin{minipage}[b]{0.32\linewidth}
	\centering
		\includegraphics[width=0.65\linewidth]{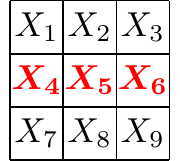}
\vspace{-0.15cm}
\caption*{top-$9$}
\end{minipage}	
\caption{\textbf{Top correlated sets discovered on Tic-tac-toe}. Color indicates the selected cells, with red designating the inclusion of $X_{10}$ that corresponds to the binary outcome of the game. In a nutshell, red and blue correlated sets can be interpreted as latent factors for win and loss, respectively.~(Sec.\ref{sec:exam})}
\label{fig:tic}
\end{figure}

Although theoretically sound, in practice normalized total correlation is \textit{unreliable} when we estimate it from empirical data: due to sparsity the plug-in estimator leads to chance-inflated estimates~\cite{romano:2016:chance}. This is particularly bad in our setting, as the data sparsity induced by the increasingly larger sets of variables we have to consider during optimization, can lead to many false discoveries (see Fig.~\ref{fig:chance} for a demonstration). Although its scores are comparable, this does not mean that normalized total correlation is easy to optimize; the score is neither monotone, nor submodular, and hence the resulting combinatorial optimization problem for discovering the top correlated sets is difficult to solve efficiently.

To address each of these issues, we build upon the recent advances on deriving corrected-for-chance information-theoretic estimators well-suited for optimization~\cite{vinh2014reconsidering,mandros2017discovering},  and propose a \textit{reliable} and \textit{efficient} estimator for normalized total correlation. The performance of this estimator is not hindered by data sparsity. Furthermore, we enable effective exhaustive and heuristic algorithms for the discovery of the top correlated sets by exploiting various structural properties of the estimator proposed. Experimental evaluation shows that the estimator has attractive statistical properties, the algorithms proposed are indeed effective on a wide range of benchmark data, and finally, concrete findings in two example applications show that our framework discovers interesting and sensible information (see Fig.~\ref{fig:tic}). Our main contributions are the following: we
\begin{enumerate}[	label=\roman*)]
	\item propose a consistent, corrected-for-chance, and efficient estimator for the normalized total correlation (Sec.~\ref{sec:score}), 
	\item provide effective algorithms for exact, approximate, and heuristic search (Sec.~\ref{sec:opt}), and finally
	\item perform empirical evaluation on a wide range of real and synthetic datasets (Sec.~\ref{sec:eval}).
\end{enumerate}

Clearly, we are far from the first to consider mining correlated sets from categorical data. Existing methods, however, all have significant drawbacks. Many methods, are primarily defined for binary data and measure only pairwise associations with interestingess functions such as $\chi^2$~\cite{brin:1997:beyond}, all-confidence~\cite{omiecinski2003alternative}, h-confidence~\cite{xiong:2006:hyper}, or mutual information~\cite{ke:2008:cor}. By considering only pairwise associations, higher-order interactions among the features are neglected. In addition, data transformations from categorical attributes to boolean may incur information loss. Finally, such methods are parameterized with various thresholds, e.g., minimum all-confidence, leading to an uncontrollable output size, i.e., they might miss interesting correlations or receive too many. In a nutshell, we find that correlation mining methods, although relevant for their own respective applications, lack a comprehensive formalization of correlation, as well as parameter-free, single-objective optimization problems for categorical data like we propose here. Total correlation has been used in other unsupervised scenarios, such as learning latent representations~\cite{steeg:2014:corex}, measuring correlation in real-valued data~\cite{nguyen2016universal,wang2017unbiased}, and mining high order interactions~in~binary~data~\cite{zhang2008mining}. 

We start with preliminaries and problem definition in Sec.~\ref{sec:prob}, propose our estimator in Sec.~\ref{sec:score}, our algorithms in Sec.~\ref{sec:opt}, and proceed with the evaluation in Sec.~\ref{sec:eval}. We round up with a concluding discussion in Sec.~\ref{sec:conc}.

	\section{Problem Definition}\label{sec:prob}

We consider data $\D_n$ consisting of $n$ i.i.d. samples from a set of $d$ categorical random variables $\cI=\{X_1, \dots, X_d\}$, with joint distribution $p(X_1, \dots, X_d)$, domains $\domain{X_i}$, and domain sizes $\domains{X_i}=\card{\domain{X_i}}$. We are interested in discovering subsets $\cX \subseteq \cI$ in $\D_n$ that exhibit high correlation/redundancy with respect to the unsupervised information-theoretic concept of total correlation introduced by Watanabe~\cite{watanabe:1960:tc}.

The \defemph{total correlation} for a set of variables $\cX=\{X_1, \dots, X_m\}$ is defined as 
\begin{align*}
\tc(\cX)=& \sum_{X \in \cX} \Big( H(X)\Big) - H(\cX) = \sum_{i =2}^{m} \mut(\cX_{i-1};X_i) \enspace,
\end{align*}
where $\cX_i$ represents the set $\{X_j \in \cX \with j \leq i \leq m \}$, with $\cX_0$ being the empty set. Here, $H$ denotes the \defemph{Shannon entropy} defined as $H(X) = -\sum_{x \in \domain{X}} p(x)\log p(x)$ for random variable $X$, and quantifies its uncertainty in bits of information, assuming logarithm with base $2$~\cite{shannon:1948:communication}. Also, $H(X \mid Y)$ denotes the \defemph{conditional entropy} of $X$ given another random variable $Y$, i.e., $H(X \mid Y)=\sum_{y \in \domain{Y}} p(y) H(X \mid Y=y)$, and quantifies the uncertainty of $X$ conditioned on $Y$. Lastly, $\mut(X;Y)=H(X)-H(X \given Y)$ is the \defemph{mutual information}, and measures the amount of shared information between $X$ and $Y$. Total correlation can be expressed as the KL-Divergence between the joint $p(\cX)$ and the product of marginals  $\prod_{X \in \cX}p(X)$. Note that total correlation is order invariant as a function of $p$.

Essentially, total correlation is a multivariate correlation/redundancy measure quantifying the total amount of shared information in a set of random variables. It holds that $\tc(\cX) \geq 0$, with equality if and only if all variables $X \in \cX$ are statistically independent, and is monotonically increasing with the subset relation, i.e., for sets of variables $\cX$ and $\cX'$ with $\cX \subseteq \cX'$, it holds that $\tc(\cX) \leq \tc(\cX')$. 

Total correlation, however, is not suitable for comparing the degree of correlation between different sets of variables, since cardinalities, joint and marginal entropies, all vary. In addition, the monotonicity property implies that larger sets are more preferable as solutions, even in situations where $\tc(\cX')=\tc(\cX) + \epsilon$ for sets $\cX \subseteq \cX'$. This introduces redundancy and might hinder next steps of the analysis, such as visualizations. Finally, total correlation lacks an intuitive and intepretable scale, e.g., in $[0,1]$, that would facilitate the process to understand the results and reason about. These can be resolved by expressing how far the correlation in a set of variables is from the scenario of them being maximally correlated. To achieve this, we present the following proposition.

\begin{proposition}
	Given a set of variables $\cX=\{X_1,\dots, X_m\}$, we have that
	\begin{enumerate}[label=\alph*)]
		\item $\tc(\cX) \leq \sum_{X \in \cX} H(X) - \max_{X \in \cX} H(X)$,
		\item with equality $\mathit{iff}$ $\exists X_i \in \cX$ s.t., $X_j = f(X_i), \forall X_j \in \cX$. 
	\end{enumerate}
\end{proposition}
\begin{proof}
	Let us first recall a few key properties regarding Shannon entropy (e.g., \cite[Ch.~2]{cover:06:elements}). For two random variables $X,Y,$ we have that $H(X\given Y)=H(X)$, if and only if $X \perp \!\!\! \perp Y$. Moreover, $H(X\given Y)=0$ if and only if $X=f(Y)$, in the statistical sense  that for all $x \in \domain{X}$, there exists $y \in \domain{Y}$ such that $p(X=x\given Y=y)=1$. In addition, Shannon entropy has the following chain rule decomposition, $H(\cX)=\sum_{i=1}^{m} H(X_i \given \cX_{i-1})$, and is monotonically increasing with the subset relation, i.e., if $\cX \subseteq \cX'$, then $H(\cX) \leq H(\cX')$. 
	
	a) We upper-bound $\tc(\cX)$ by lower bounding $H(\cX)$. Since Shannon entropy is monotonically increasing with the subset relation, we have that $H(\cX) \geq \max_{X \in \cX} H(X)$, and hence $\tc(\cX) \leq \sum_{X \in \cX} H(X) - \max_{X \in \cX} H(X)$.
	
	b) 	Suppose that $\tc(\cX)=\sum_{X \in \cX} H(X) - \max_{X \in \cX} H(X)$. Then $H(\cX)=\max_{X \in \cX} H(X)=H(X_q)$ for some $q \in [1,m]$. Using the chain rule, and  since this decomposition is order-invariant, it is clear  that $H(X_i \given X_q)=0$ for all $X_i \in \cX$. This is possible if and only if $X_i=f(X_q)$ for all $X_i \in \cX$.
	
	Conversely, suppose there exists $X_q \in \cX$ s.t., $X_j = f(X_q), \forall X_j \in \cX$. Hence, we have that $H(X_j \given X_q)=0$ for all $X_j \in \cX$, and $H(\cX)=H(X_q)$. Now,  $X_q=\max_{X \in \cX} H(X)$, i.e., $X_q$ must be the variable with the highest entropy, hence $\tc(\cX)=\sum_{X \in \cX} H(X) - \max_{X \in \cX} H(X)$.
\end{proof}

We now define $\tcbound(\cX)=\sum_{X \in \cX} H(X) - \max_{X \in \cX} H(X)$, and proceed to define the \defemph{normalized total correlation} as
\begin{equation}
\ntc(\cX)=\tc(\cX)/\tcbound(\cX) \enspace,
\end{equation}
for which it holds that $\ntc(\cX)\in[0,1]$, with $0$ being the case where all $X \in \cX$ are statistically independent, and $1$ when there exists a variable that ``explains" all others.\!\footnote{note that the bound for total correlation is in general known in the literature, e.g.,~\cite{watanabe:1960:tc}. However, a formal proof for the bound is often missing, which we present here for both self-containment, and to better understand its properties.} By quantifying the percentage of correlation within $\cX$, the score is now better interpretable, as well as comparable across the different variable sets with varying joint and marginal entropies.

\begin{figure}[t]
		\centering
			\includegraphics[scale=1]{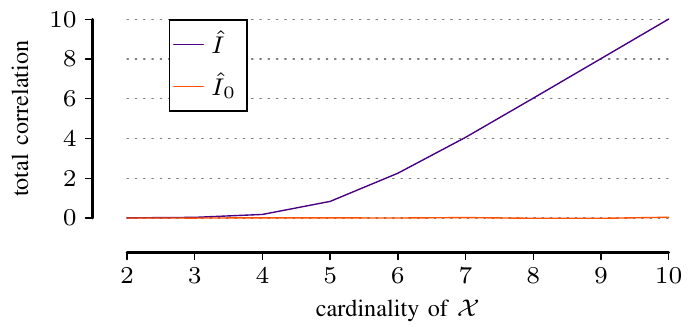}
			\caption{\textbf{Correlation-by-chance.} 
				Estimated total correlation for variable set $\cX$ of increasing cardinality. All variables are uniformly and independently sampled with domain size $4$ and sample size $1000$. Population value for total correlation is $0$. Correlation increases when naive estimator $\hmut$ is used, but not for the corrected-for-chance $\hmuto$. (\cite{mandros2017discovering}, also defined in Sec.~\ref{sec:score})}
		\label{fig:chance}
\end{figure}

The data $\D_n$ induce an empirical distribution $\hatp$  defined using the empirical counts of values in $\D_n$, from which plug-in estimators can be derived for all the aforementioned quantities, i.e., $\hH,\hmut, \htc, \hntc$. These estimators, however, are known to have biases that depend on the domain sizes of the variables involved~\cite{roulston1999estimating}, with mutual information, in particular, having a positive bias. While it is easier in general to obtain good estimates for marginal quantities, total correlation involves mutual information terms that need to be estimated for increasingly larger sets of variables. This can lead to situations with arbitrary estimates (see Fig.~\ref{fig:chance} for a demonstration). 

Even if a more suitable estimator was available, the resulting combinatorial optimization problem for finding  the top correlated sets $\cX^*$ in $\D_n$ is in practice infeasible for naive solutions. Hence, in order to have an overall useful method for our task, we need to a) derive a corrected-for-chance estimator $\hntc'$ for $\ntc$, and b) find an effective solution to the optimization problem by exploiting structural properties of $\hntc'$. We present solutions to these  in Sec.~\ref{sec:score} and Sec.~\ref{sec:opt} respectively.

	\section{Reliable normalized total correlation}\label{sec:score}

In this section we derive a corrected for chance, consistent, and efficient to compute estimator for the normalized total correlation. The estimator follows the idea of correcting the plug-in by subtracting values of suited null hypothesis models, leading to either  parametric (e.g., \cite{vinh2014reconsidering}), or non-parametric solutions  (e.g., \cite{nguyen:2010:chancejournal}). Unlike the plug-in, such estimators give conservative estimates for sparse data in high-dimensional spaces, making them therefore well-suited for reliable optimization.

For the non-parametric case, Mandros et al.~\cite{mandros2017discovering} propose an estimator for mutual information defined as
\begin{equation}
\hmuto(X;Y)= \hmut(X;Y)-\Eo{\hmut(X;Y)} \enspace,
\end{equation}
where $\Eo{\hmut(\cX;Y)}$ is the expected value of $\hmut$ under the \defemph{permutation model}~\cite[p. 214]{lancaster:1969:chi}, a non-parametric independence model for contingency tables that assumes fixed marginal counts. The expected value under this model is equal to $\Eo{\hmut(\cX;Y)}=\nicefrac{ \sum_{\sigma \in S_n} \hmut(X;Y_\sigma)}{n!}$, where $S_n$ denotes the symmetric group for $n$, i.e., the set of all bijections from $\{1,\dots,n\}$ to $\{1,\dots,n\}$, and $Y_\sigma$ denotes the $Y$ samples permuted according to a $\sigma \in S_n$. Exploiting symmetries, this value can be computed in $O(n\max\{\domains{X},\domains{Y}\})$ (see \cite{vinh2009information,mandros2018discovering} for the computation, and \cite{romano:2014:smi} for the complexity). For the rest of this paper we denote $\Eo{\hmut(X;Y)}$ with $\mo(X,Y,n)$.

Following the same non-parametric correction principle, and assuming we can adequately estimate marginal entropies $\hH(X)$, we can define a corrected-for-chance estimator for the normalized total correlation by plugging $\hmuto$ and arrive at
\begin{align}
\sum_{i=2}^{m} \Big( \hmut(\cX_{i-1};X_i) - \mo(\cX_{i-1},X_i,n) \Big)/\tcbound(\cX)  \enspace.
\end{align}
However, unlike the plug-in $\hntc$, this estimator violates the order-invariance of total correlation since the correction $\mo$ is not a function of $\hatp$, but rather a function of domain sizes and marginal counts. To ensure order-invariance, we select the order of variables that leads to the most conservative estimate for the normalized total correlation, which translates to the order that maximizes the correction term, i.e.,
\begin{align}
\hntco(\cX)=&\frac{\sum_{i=2}^{m} \hmut(\cX_{i-1};X_i)}{\tcbound(\cX)} \\
&- \frac{\max \limits_{\sigma \in S_m} \sum_{i=2}^{m} \mo(\cX_{\sigma(i-1)},X_{\sigma(i)},n)}{\tcbound(\cX)} \\
=& \hntc(\cX)-  \tzero(\cX,n)   \enspace,
\end{align}
where $\cX_\sigma$ denotes set $\cX$ ordered according to a $\sigma \in S_m$.

Regarding efficiency, $\hntco$ is clearly infeasible to compute in practice. For a set of $m$ variables, there are $m-1$ calculations of the permutation model with each subsequent calculation having an increased cost (since domain sizes $\domains{\cX_{\sigma(i-1)}}$ can grow exponentially with $i$), and there are $m!$ possible permutations to find the maximum correction term, resulting in a total complexity of $O(m^2(m-1)!n\domains{\cX})$. We dramatically reduce this complexity by first replacing the exact calculation of the expected value $\mo$ with an upper bound, and then propose a relaxation to this bound such that we can efficiently find the order $\sigma^* \in S_m$ of variables maximizing the correction term.  

\begin{proposition}[\cite{nguyen:2010:chancejournal}, Thm.~7]
	\label{prop:upperbound}
	For variables $X,Y,$ with domain sizes $\domains{X}, \domains{Y},$ and sample size $n$, it holds that
	\begin{align*}
		\mo(X,Y,n) &\leq \log\frac{n+\domains{X}\domains{Y}-\domains{X}-\domains{Y}}{n-1} \enspace.
	\end{align*}
\end{proposition}


We denote this upper bound with $\mbo(X,Y,n)$, and the corresponding correction term $\tbzero(\cX,n)$, i.e.,
\begin{equation}
\tbzero(\cX,n)=	\max \limits_{\sigma \in S_m} \sum_{i=2}^{m} \mbo(\cX_{\sigma(i-1)},X_{\sigma(i)},n)/\tcbound(\cX) \enspace.
\end{equation}

Now, while the exact expected values have been replaced with something more efficient, $\tbzero(\cX,n)$  as function of  the joint domain sizes $\domains{\cX_{\sigma(i-1)}}$ remains infeasible: for every $\sigma \in S_m$ and $i\in[2,m]$, we need to compute the joint domain size of $\cX_{\sigma(i-1)}$ with $X_{\sigma(i)}$. We proceed to relax this requirement.

Assuming a strictly positive distribution $p$, i.e., $p(\cX=\mathbf{x}) > 0$ for all $\cX\subseteq \cI$ and $\mathbf{x}\in \domain{\cX}$,  then joint domain sizes can be written as a product of marginal domain sizes, i.e., $\domains{\cX}=\prod_{X \in \cX} \domains{X}$. Furthermore, a relaxation that considers only the joint contribution of the variables in $\cX$, leads to the  bound 

\begin{align*}
\mbop(\cX_{i-1},X_i,n)=\log\frac{n+\big(\prod_{X \in \cX_{i-1}}\domains{X}\big)\domains{X_i}}{n-1} \enspace,
\end{align*}
 and to the following correction term
\begin{equation}
\tbzerop(\cX,n)=\max \limits_{\sigma \in S_m} \sum_{i=2}^{m} \mbop(\cX_{\sigma(i-1)},X_{\sigma(i)},n)/\tcbound(\cX) \enspace.
\end{equation}


In the following theorem we establish that this quantity is both a consistent upper bound for $\tbzerop$, and efficient to compute without explicitly considering all permutations $\sigma \in S_m$.

%

\begin{theorem}\label{th:eff}
For set of variables $\cX=\{X_1, \dots, X_m\}$, it holds 
	\begin{enumerate}[label=\alph*)]
	\item $\tbzerop(\cX,n) \geq \tbzero(\cX,n) $
	\item\label{th:cons}$\lim_{n \rightarrow \infty} \tbzerop(\cX,n)=0$
	\item $\sum_{i=2}^{m} \mbop(\cX_{\sigma(i-1)},X_{\sigma(i)},n)$ is maximized for $\sigma^* \in S_m$ with $\domains{X_{\sigma^*(1)}} \geq \domains{X_{\sigma^*(2)}} \dots \geq \domains{X_{\sigma^*(m)}}$

\end{enumerate}
\end{theorem}
\begin{proof}
For readability, we drop $\sigma$ as a subscript whenever clear from the context. \\
	a) We prove this statement by first showing that it holds for any $\sigma \in S_m$. Given a $\sigma \in S_m$, and any $i \in [2,m]$, we have 
	\begin{align*}
		\mbop(\cX_{i-1},X_i,n)=& \log \frac{n+\domains{X_i} \prod_{X \in \cX_{i-1}} \domains{X}}{n-1}  \\
		\geq& \log \frac{n+\domains{X_i} \!\!\! \prod \limits_{X \in \cX_{i-1}} \!\!\! \domains{X} - \prod \limits_{X \in \cX_{i-1}} \!\!\! \domains{X} -\domains{X_i} }{n-1}    \\
		\geq&\log \frac{n+\domains{\cX_{i-1}} (\domains{X_i}-1)-\domains{X_i}}{n-1} \\
		=&\mbo(\cX_{i-1},X_i,n)	\enspace,
	\end{align*}
since $\prod_{X \in \cX_{i-1}}\domains{X} \geq \domains{\cX_{i-1}}$ and $\log$ is a monotonically increasing function. Because this holds for any $\sigma \in S_m$ and $i \in [2,m]$, then for the $\sigma^*$ with $\sigma^*=\argmax_{\sigma \in S_m} \sum_{i=2}^{m} \mbo(\cX_{\sigma(i-1)},X_{\sigma(i)},n)$ we have that $\sum_{i=2}^{m} \mbop(\cX_{\sigma^*(i-1)}, X_{\sigma^*(i)},n)$ is larger. 

b) It follows from $\lim_{n \rightarrow \infty} \log(\frac{(n+a)}{(n-1)})=0$.

c) Let us consider a $\sigma^* \in S_m$ for which $\domains{X_{\sigma^*(1)}}
 \geq \dots  \geq \domains{X_{\sigma^*(m)}}$, and any arbitrary $\sigma \in S_m$. We prove this statement by doing a pairwise comparison between $\mbop(\cX_{\sigma{(i-1)}},X_{\sigma(i)},n)$ and $\mbop(\cX_{\sigma^*{(i-1)}},X_{\sigma^*(i)},n)$ for any $i \in [2,m]$.  We have
\begin{align*}
\mbop(\cX_{\sigma^*(i-1)},X_{\sigma^*(i)},n)=& \log \frac{n+\prod_{X \in \cX_{\sigma^*(i)}} \domains{X}}{n-1} \\
\geq& \log \frac{n+\prod_{X \in \cX_{\sigma(i)}} \domains{X}}{n-1}\\
=&\mbop(\cX_{\sigma(i-1)},X_{\sigma(i)},n) \enspace,
\end{align*}
where the inequality follows from the fact that $\prod_{X \in \cX_{\sigma^*(i)}}\domains{X}$ is the product of the $i$ largest domain sizes. Since this holds for any $\sigma \in S_m$ and $i \in [2,m]$, then $\sigma^*=\argmax_{\sigma \in S_m} \sum_{i=2}^{m} \mbop(\cX_{\sigma(i-1)},X_{\sigma(i)},n)$.
\end{proof}

We now have an efficiently computable correction term $\tbzerop(\cX,n)$, going from an initial complexity of $O(m^2(m-1)!n\domains{\cX})$, to that of $O(m+m\log m)$, where $m\log m$ is for sorting the domain sizes $\domains{X}$, for $X \in \cX$. In addition, as an upper bound to $\tbzero$,  this correction is as conservative with regards to its estimates, which is a design goal for reliability. Finally, we arrive at the \defemph{reliable normalized total correlation}

\begin{equation}
\hntcbd(\cX)=\hntc(\cX)-\tbzerop(\cX,n) \enspace.
\end{equation}
In addition to being very efficient, the consistency of the plug-in $\hH$ (see, e.g., \cite{antos2001convergence}), together with Th.~\ref{th:eff}\ref{th:cons}, implies that $\hntcbd$ is a consistent estimator for the normalized total correlation. 

The estimators discussed here are evaluated further for their statistical properties in Sec.~\ref{sec:stat}.

	\section{Optimization}\label{sec:opt}

Here, we provide algorithms for the following optimization problem: given data $D_n$ consisting of $n$ i.i.d. samples of random variables $\cI=\{X_1, \dots, X_d\}$, as well as a positive integer $k$, find the top-$k$ subsets $\cX^*_{1}, \dots, \cX^*_{k} \subseteq \cI$ with
\begin{equation}\label{eq:problemEst}
\hntcbd(\cX^*_{i}) =\max\{ \hntcbd(\cX) \with \hntcbd(\cX^*_{i-1}) \geq \hntcbd(\cX), \cX \subseteq \cI\} \enspace.
\end{equation}

Given the combinatorial nature of the problem, as well as the recent hardness result for optimizing the reliable mutual information $\hmuto$~\cite{mandros2018discovering}, it seems unlikely that the optimization of $\hntcbd$ allows for a polynomial algorithm. While the complexity of the optimization problem under consideration is an open question, here we derive two practically efficient algorithms for exhaustive and heuristic search. 

As is common in hard combinatorial problems, we instantiate the exact algorithm with the branch-and-bound framework (see, e.g., \cite[Chap. 12.4]{mehlhorn2008algorithms}). To recall the basics, branch-and-bound, as the name suggests, consists of two main ingredients: a strategy to enumerate some abstract search space $\Omega$,  and an admissible bound for the optimization function $\mapping{f}{\Omega}{\mathbb{R}}$ at hand. The former is governed by the \defemph{branch operator}, a function $\mapping{\branch}{\mathcal{P}(\Omega)}{\mathcal{P}(\Omega)}$ that non-redundantly generates the search space from some designated root element $\bot \in \Omega$, i.e., for all $\omega \in \Omega$ there must be a unique sequence $\bot=\omega_1,\dots,\omega_l=\omega$ such that $\omega_{i+1} \in \branch(\omega_i)$ for $i=1,\dots,l-1$. 

An \defemph{admissible bounding function} $\oest$, also known as optimistic estimator, must guarantee the property $\oest(\omega) \geq \max \lbrace f(\omega') \with \omega' \in \branch^*(\omega) \rbrace$, where $\branch^*(\omega)$ denotes the set of all $\omega' \in \Omega$ that can be generated from $\omega$ by multiple applications of $\branch$. The value $\oest(\omega) $ is called the \textbf{potential} of element $\omega$.
With these, a branch-and-bound algorithm enumerates $\Omega$ starting from $\bot$, tracks the best solution, and prunes expanding elements with $\oest$ that cannot yield an improvement over the best solution. In addition, the framework provides the option of relaxing the required result guarantee to that of an $\alpha$-approximation for accuracy parameter $\alpha \in (0,1]$. Therefore, an  $\alpha <1$ allows to trade accuracy for efficiency in a principled manner. 

Essentially a bounding function is a worse case scenario for the maximum attainable score $\hntcbd(\cX')$ for supersets of $\cX$ in the enumerated search space. Hence, the ideal one would be 
\begin{equation}
\bar{w}^*_{\bar{\bar{0}}}(\cX)=\max\{\hntcbd(\cX') \with \cX \subseteq \cX' \subseteq \cI \} \enspace.
\end{equation}
Efficiently computing this function, however, would imply an efficient algorithm for the original optimization problem. Instead, we shift our attention into independently deriving tight bounds for the two terms of $\hntcbd(\cX)$, i.e., an upper bound for $\hntc(\cX)$ and a lower bound for $\tbzerop(\cX,n)$, in order to arrive at a looser, but efficient to compute bounding function. In our setting, however, it is not possible to both derive tight bounds and also guarantee their admissibility for arbitrarily enumerated search spaces. The difficulty stems from the inability to ``predict'' their behavior with respect to the subset relation---both numerators are monotonically increasing functions, but this property does not extend together with the normalizer $\tcbound(\cX)$. For example, for a $\cX' \supseteq \cX$ it might be that $\tbzerop(\cX',n) \geq \tbzerop(\cX,n) $, but for a different superset $\cX'' \supseteq \cX$ that $\tbzerop(\cX'',n) \leq \tbzerop(\cX,n)$. In other words, anything can happen. 

As it turns out, under a more strict partial order we can induce a certain structure into our problem that allow us to derive tight, admissible bounds for both terms. 
\begin{definition}
	Given $\cI=\{X_1, \dots, X_d\}$, we say that $\cX' \subseteq \cI$ is a \defemph{low entropy extension} of a $\cX \subseteq \cI$, denoted as $\cX \lmer \cX'$, if $\cX \subseteq \cX'$, and for all $X' \in \cX'\setminus \cX$, $ \hH(X') \leq   \min_{X \in \cX} \hH(X)$.
\end{definition}

We can guarantee that this partial order holds in the enumerated search space by simply considering a decreasing-entropy branching operator of the form
\begin{equation}
\branch_{H}(\cX)=\{\cX \cup \{X\} \with \hH(X) \leq \min_{X' \in \cX} \hH(X'), X \in \cI \setminus \cX \} \enspace,
\end{equation}
i.e., it holds that $\cX \lmer \cX'$ for all $\cX' \in \branch_{H}(\cX)$. We now proceed with showing that under this partial order, the correction term $\tbzerop$ is monotonically increasing. First, we provide the following required Lemma.

\begin{lemma}\label{lem:frac}
	For two fractions $\nicefrac{a}{x}$ and $\nicefrac{b}{y}$ of positive integers, if $\nicefrac{a}{x} \leq \nicefrac{b}{y}$, then it holds that $\nicefrac{a}{x} \leq \nicefrac{(a+b)}{(x+y)}$.
\end{lemma}
\begin{proof}
	We have
	\begin{align}
	\frac{a}{x} \leq \frac{b}{y} \Rightarrow	 ay \leq  bx &\Rightarrow ay +ax \leq  bx +ax \Rightarrow	 \\ 
	\frac{ay +ax}{x(x+y)} \leq  \frac{ax +bx}{x(x+y)} &\Rightarrow \frac{a(y +x)}{x(x+y)} \leq  \frac{x(a +b)}{x(x+y)}	 \Rightarrow \\
	\frac{a}{x} &\leq  \frac{a+b}{x+y} \enspace,
	\end{align}
	concluding the proof.
\end{proof}

\begin{theorem}\label{th:mon}
	For subsets $\cX,\cX'$ of $\cI$ with $\cX \lmer \cX'$, it holds that  $\tbzerop(\cX,n) \leq \tbzerop(\cX',n)$.
\end{theorem}
\begin{proof}
 Let $\cX=\{X_1, \dots, X_m\}$ and $\cX'=\cX \cup \cZ$, with $\cZ=\{Z_1, \dots, Z_q\}$. Let us assume for simplicity and w.l.o.g. that $X_1$ is the variable with the maximum entropy in $\cX$, and that $\domains{X_1}\geq \dots \geq\ \domains{X_d}$ and $\domains{Z_1} \geq \dots \geq \domains{Z_q}$.\!\footnote{the former allows us to write the normalizer $\tcbound(\cX)$ as $\sum_{i=2}^{m} \hH(X_i)$, and the latter to remove the $\max$ operator from the numerator of $\tbzerop$.} In addition, let us assume for now that $\min_{X \in \cX} \domains{X} \geq \max_{Z \in \cZ} \domains{Z}$.
 
 Since $\cX \lmer \cX'$, $X_1$ is also the largest entropic variable in $\cX'$, and because  $\min_{X \in \cX} \domains{X} \geq \max_{Z \in \cZ} \domains{Z}$, we can separate the contributions of $\cX$ and $\cZ$ and reformulate $\tbzerop(\cX',n)$ as
\begin{align*}
	\frac{\sum_{i=2}^{m} \mbop(\cX_{i-1},X_i,n) + \sum_{j=1}^{q} \mbop(\cX \cup \cZ_{j-1},Z_j,n)  }{\sum_{i=2}^{m} \hH(X_i) + \sum_{j=1}^{q}\hH(Z_j)} \enspace.
\end{align*}
Now let us use the notation $a=\sum_{i=2}^{m} \mbop(\cX_{i-1},X_i,n)$, $b=\sum_{j=1}^{q} \mbop(\cX \cup \cZ_{j-1},Z_j,n)$, $x=\sum_{i=2}^{m} \hH(X_i)$, $y=\sum_{j=1}^{q}\hH(Z_j)$. We need to show that $\frac{a+b}{x+y} \geq  \frac{a}{x}$.  

As $\cX \lmer \cX'$, we have that $\sum_{j=1}^{q}\hH(Z_j)$ is a sum of $q$ terms, smaller than the $m-1$ terms of  $\sum_{i=2}^{m}\hH(X_i)$. In addition, and by the definition of $\mbop$, the quantity $\sum_{j=1}^{q} \mbop(\cX \cup \cZ_{i-1},Z_i,n) $ is a sum of $q$ terms larger than the $m-1$  terms of $\sum_{i=2}^{m} \mbop(\cX_{i-1},X_i,n)$. Hence, the fraction $b/y$ is larger than that of $a/x$, and from Lem.~\ref{lem:frac}, we have that $\frac{a+b}{x+y} \geq  \frac{a}{x}$.

 Now if it were not the case that $\min_{X \in \cX} \domains{X} \geq \max_{Z \in \cZ} \domains{Z}$, i.e., there exist variables in $\cZ$ with domain sizes larger than those in $\cX$, then we could still write the numerator of $\tbzerop(\cX')$ as two sums $a'$ and $b'$ with $m-1$ and $q$ terms respectively, and it would hold that $a' \geq a$ and $b' \geq b$, and hence
\begin{align*}
\frac{a}{x} \leq \frac{a+b}{x+y} \leq \frac{a'+b'}{x+y} \enspace,
\end{align*} 
concluding the proof.
\end{proof}

Following from the theorem, a \textbf{trivial bounding function} can be derived using the upper bound $1$ for $\hntc(\cX)$, i.e.,
\begin{align*}
\hntcbd(\cX') = & \hntc(\cX') -\tbzerop(\cX',n) \\   
\leq & 1-\tbzerop(\cX,n)=\btcmon(\cX) \enspace,
\end{align*}
for all $\cX'$ that are low entropy extensions of $\cX$. It is clear, however, that $\btcmon(\cX)$ is not tight: it upper bounds $\hntc(\cX)$ with the maximum possible value for the normalized total correlation, without taking into consideration both the correlation in $\cX$ so far, nor how ``good" it might actually become for $\cX'$. We derive a much tighter upper bound for $\hntc$ by further exploiting the structure of the enumerated space. We define $R_{\cX}=\{X \with \hH(X) \leq \min_{X' \in \cX} \hH(X'), X \in \cI \setminus \cX  \}$ to be the set of all refinement elements of $\cX$, and $\btc(\cX)$ the quantity
\begin{equation}
	\btc(\cX)=\frac{\sum_{i=2}^{m} \hmut(\cX_{i-1};X_i) + \sum_{X' \in R_{\cX}} \hH(X')}{\tcbound(\cX) + \sum_{X' \in R_{\cX}} \hH(X')} \enspace,
\end{equation} 
i.e., the plug-in $\hntc(\cX)$ after adding the marginal entropies of the refinement elements of $\cX$. The following theorem establishes that $\btc(\cX)$ is an upper bound to $\hntc(\cX)$ with respect  to $\lmer$.
\begin{theorem}\label{th:bound}
	For a $\cX \subseteq \cI$ and any $\cX' \subseteq \cI$ with $\cX \lmer \cX'$, it holds that $\btc(\cX) \geq \hntc(\cX')$.
\end{theorem}
\begin{proof}
	Let $\cX=\{X_1, \dots, X_m\}$ and $\cX'=\cX \cup \cZ$, with $\cZ=\{Z_1, \dots, Z_q\}$. We have
	\begin{align}
		\hntc(\cX')&=\frac{\sum_{i=2}^{m} \hmut(\cX_{i-1};X_i) + \sum_{j=1}^{q} \hmut(\cX \cup \cZ_{j-1};Z_j)}{\tcbound(\cX) + \sum_{j=1}^{q} \hH(Z_j)} \\
		&\leq \frac{\sum_{i=2}^{m} \hmut(\cX_{i-1};X_i) + \sum_{j=1}^{q} \hH(Z_j)}{\tcbound(\cX) + \sum_{j=1}^{q} \hH(Z_j)} \\
		&\leq \frac{\sum_{i=2}^{m} \hmut(\cX_{i-1};X_i) + \sum_{j=1}^{q} \hH(Z_j) + \!\!\!\!\!\! \sum \limits_{X' \in R_{\cX'}} \!\!\!\! \hH(X') }{\tcbound(\cX) + \sum_{j=1}^{q} \hH(Z_j) + \sum_{X' \in R_{\cX'}} \hH(X')} \\
		&= \frac{\sum_{i=2}^{m} \hmut(\cX_{i-1};X_i) + \sum_{X' \in R_{\cX}} \hH(X')}{\tcbound(\cX) + \sum_{X' \in R_{\cX}} \hH(X')} = \btc(\cX) \enspace,
	\end{align}
	where the first inequality follows from the fact that $\hmut(X;Y) \leq \min\{\hH(X),\hH(Y)\}$ for variables $X$ and $Y$~\cite[Ch.~2]{cover:06:elements}, and that $\cX \lmer \cX'$, i.e., $\hmut(\cX \cup \cZ_{j-1};Z_j) \leq \hH(Z_j)$ for all $j \in [1,q]$. The second inequality follows from Lem.~\ref{lem:frac}.
\end{proof}

We can now define the \textbf{tighter bounding function} $\btcref(\cX)=\btc(\cX)-\tbzerop(\cX,n)$, which has an extra $O(\card{\cR_{\cX}})$ complexity compared to $\btcmon(\cX)$. Note that in practice we use both: first evaluate $\btcmon$ that we get for free by caching $\tbzerop$ after computing $\hntcbd$, and then proceed with $\btcref$ if it fails.

The pseudocode in Algorithm~\ref{alg:bnb} summarizes the resulting \textbf{exhaustive} method for the discovery of reliable correlated sets. For simplicity, we present the top-$1$ formulation. The algorithm maintains a priority queue $\mathbf{Q}$ that holds the search frontier and a current result set $\cS$ throughout the search. As long as the queue is not empty, the search continuous by expanding the top element (line 5), updating the current result set (line 6), pruning (line 7), and updating the queue (line 9). For heuristic search, we consider the standard \textbf{greedy} algorithm, i.e., level-wise search where only the best candidate is refined, coupled with $\branch_{H}$ and $\btcref$ for pruning. 

Regarding practicalities, for branch-and-bound we use a priority queue based on potential that leads to the best-first variant. The branching operator $\branch_{H}$  is equivalent to the standard alphabetical enumeration with $\branch(\cX)=\{\cX \cup \{X_i\} \with i > \max\{j \with X_j \in \cX\}, i \leq d\}$ after initially sorting the input variables in decreasing entropy order. Since $\ntc(\cX)$ is undefined for $\card{\cX}\leq1$, we define potential $1$ for $\card{\cX}=1$, and a score of $0$ for $\card{\cX}\leq1$. Moreover, the enumeration order allows for an efficient incremental calculation of $\hntcbd$.

\begin{algorithm}[t]
	\caption{
		$\bnb$: Given a set of input variables $\cI$, function $\hntcbd$, bounding function $\btcref$, branching operator $\branch_{H}$, and $\alpha \in (0,1]$,  the algorithm returns the $\cX^* \subseteq \cI$ satisfying $\hntcbd(\cX^*) \geq \alpha \max \{\hntcbd(\cX') \with \cX' \subseteq \cI \}$
	}
	\label{alg:bnb}
	\begin{algorithmic}[1]
		\Function{$\bnb$}{$\mathbf{Q},\cS$} 
		\If{$\mathbf{Q}$ is empty}
		\State \textbf{return} $\cS$
		\Else
		\State $\mathbf{R}= \branch_{H}(\text{top} (\mathbf{Q}))$
		\State $\cX^* = \argmax    \{\hntcbd(\cX') \with \cX' \in \mathbf{R} \cup \{\cS\} \} $
		\State $\mathbf{R}'=\{\cX' \in \mathbf{R}  \with \alpha \btcref(\cX') > \hntcbd(\cX^*)   \}  $
		\State $\mathbf{Q}'=(\mathbf{Q}\setminus top(\mathbf{Q}) ) \cup \mathbf{R}'$
		\State \textbf{return} $\bnb(\mathbf{Q}',\cX^*) $
		\EndIf
		\EndFunction
		\State $\cX^*=\bnb(\{ \emptyset \},\emptyset) $
	\end{algorithmic}
\end{algorithm}
	\section{Evaluation}\label{sec:eval}

In this section we empirically evaluate the proposed discovery framework for correlated patterns. In particular, we perform experiments on synthetic data in order to investigate the performance of the estimators, we use a wide selection of benchmark data to evaluate the performance of the algorithms and bounding function $\bhntcd$, as well as provide concrete findings in example exploratory tasks.

\subsection{Estimator performance}\label{sec:stat}
\begin{figure}[t]		
	\centering
				\includegraphics[scale=1]{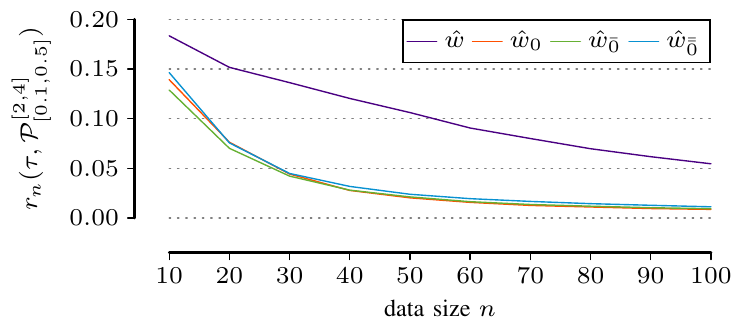}
	\caption{\textbf{Average regret.}  Regret $r_n(\tau, \cP^{[2,4]}_{[0.1,0.5]})$ for sample sizes $n=\{10,\dots, 100\}$ and estimators $\tau=\{\hntc, \hntco,\hntcbo,\hntcbd\}$.}
	\label{fig:conAll}
\end{figure}

Here we evaluate the performance of the estimators discussed in this paper, i.e., the corrected-for-chance $\hntco,\hntcbo,\hntcbd$ proposed, and the plug-in $\hntc$. For this evaluation, we first create synthetic data in the following way. We randomly and uniformly sample joint probability distributions $p^{(i)} \in \cP^{d}_{[a,b]}$, where $\cP^{d}_{[a,b]}$ denotes the set of all joint probability distributions with $d$ dependent random variables and resulting $\ntc$ score in $[a,b]$. Each random variable has a domain size of $3$. For example, $\cP^{4}_{[0,0.3]}$ is the set of probability distributions $p(\cX)$, $\cX=\{X_1,\dots,X_4\}$, with $\domains{X_i}=3$, and $\ntc(\cX) \in [0,0.3]$. We augment these distributions with $3$ independent and uniformly distributed random variables, also of domain size $3$. Each $p^{(i)} \in \cP^{d}_{[a,b]}$ has then its own set of $2^{d+3}-1$ marginalized distributions for which we can compute the $\ntc$ score. Note that due to the varying marginal entropies $H$ of the normalizer, it is not guaranteed that the full (original) joint has the highest $\ntc$, but rather that the maximum is at least as large. 

\begin{figure*}[t]
	\centering
	\begin{minipage}[b]{0.32\linewidth}
		\centering
				\includegraphics[scale=1]{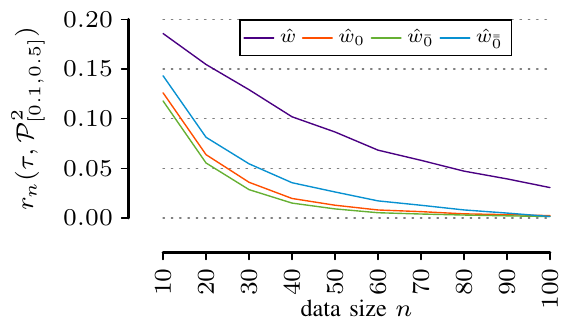}
	\end{minipage}
	\begin{minipage}[b]{0.32\linewidth}
		\centering
					\includegraphics[scale=1]{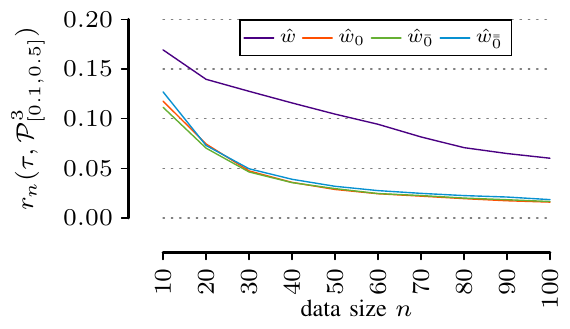}
	\end{minipage}
	\begin{minipage}[b]{0.32\linewidth}
		\centering
					\includegraphics[scale=1]{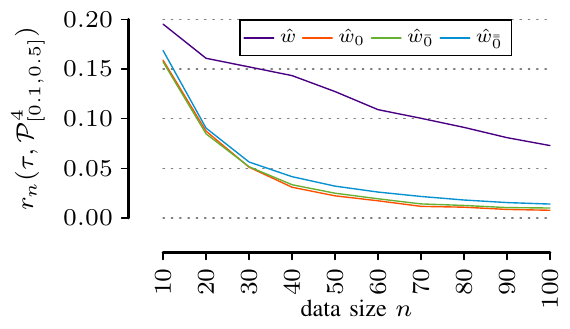}
	\end{minipage}
	\caption{\textbf{Regret curves averaged over different dimensionalities.} Average regret $r_n(\tau, \cP^{2}_{[0.1,0.5]})$ (\textbf{left}), $r_n(\tau, \cP^{3}_{[0.1,0.5]})$ (\textbf{middle}), and $r_n(\tau, \cP^{4}_{[0.1,0.5]})$ (\textbf{right}), for sample sizes $n=\{10,\dots, 100\}$ and estimators $\tau=\{\hntc, \hntco,\hntcbo,\hntcbd\}$.}
	\label{fig:conLevels}
\end{figure*}

\begin{figure}[t]
	\centering
	\begin{minipage}[b]{0.475\columnwidth}
		\centering
						\includegraphics[scale=1]{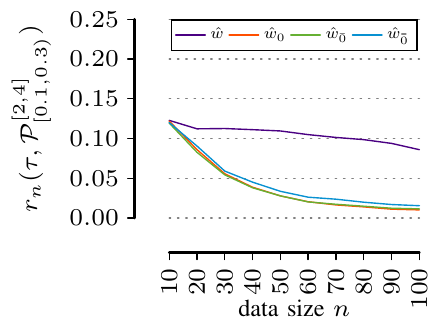}
	\end{minipage}
	\begin{minipage}[b]{0.475\columnwidth}
		\centering
							\includegraphics[scale=1]{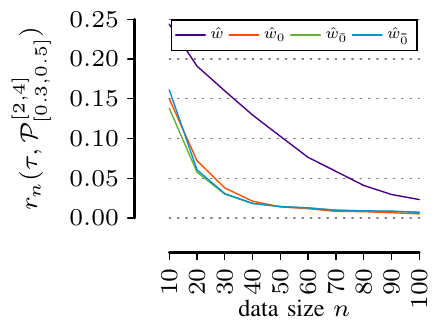}
	\end{minipage}
	\caption{\textbf{Regret curves averaged over ``low" and ``high" correlation.} Average regret $r_n(\tau, \cP^{[2,4]}_{[0.1,0.3)})$ (\textbf{left}), and $r_n(\tau, \cP^{[2,4]}_{[0.3,0.5]})$ (\textbf{right}), for sample sizes $n=\{10,\dots, 100\}$ and estimators $\tau=\{\hntc, \hntco,\hntcbo,\hntcbd\}$.}
	\label{fig:conStrengths}
\end{figure}

We consider dimensionalities $d=2,3,4,$ and four different regimes $P^d_{[0.1, 0.2)}$, $P^d_{[0.2, 0.3)}$, $P^d_{[0.3, 0.4)}$, $P^d_{[0.4, 0.5]}$, representing weak, low, medium, and high correlation.\!\footnote{note that randomly sampling joint distributions with high normalized total correlation, e.g., in $[0.5,1]$, is in practice hard for increasing dimensionalities since it requires that all conditional distributions are highly peaked. In addition, this range is less challenging for estimators as it is easily separated from noise.} We sample one distribution for each combination, resulting in $12$ different distributions $p^{(i)},i=1,\dots,12$. We consider data sizes $n=\{10,20,30,\dots, 100\}$, and for each $p^{(i)}$ and $n$ we sample $500$ datasets according to $p^{(i)}$ and denote them as $\D_{n,j}^{(i)}, j \in [1,500]$. We pick $n =\{10,\dots,100\}$, since the probability distributions we consider are ``small" in size. It is expected, given that all estimators are consistent, that their behavior carries on for larger sample sizes and distributions.

We choose regret to evaluate the estimators as it is an accurate summary of essential properties for an estimator, such as consistency, convergence, and generalization error. The \textbf{regret} is defined as $r_n(\tau, p^{(i)})=\expected{\ntc(\cX^*_i)-\ntc(\cX^*_{i,j,n,\tau})}$, where  $\cX^*_i$ represents the true maximizer of population $p^{(i)}$, and $\cX^*_{i,j,n,\tau}$ the maximizer in $\D_{n,j}^{(i)}$ according to an estimator $\tau=\{\hntc, \hntco, \hntcbo, \hntcbd\}$, for which we use exhaustive search to obtain.\!\footnote{the $d+3$ variables are the input variables, the rows are the samples, and an estimator is used as the function to be optimized} The expected value is with respect to $j\in[1,500]$. We average regrets across the different $p^{(i)}$ to obtain $r_n(\tau, \cP^{[u,v]}_{[a,b]})$, e.g., $r_n(\tau, \cP^{[2,3]}_{[0,0.5]})$ would be the average regret of estimator $\tau$ across all $p^{(i)} \in \cP^{3}_{[0,0.5]}$ and  $p^{(i)} \in \cP^{4}_{[0,0.5]}$.

We start with Fig.~\ref{fig:conAll} and plot $r_n(\tau, \cP^{[2,4]}_{[0.1,0.5]})$, i.e. the average regret across all $p^{(i)}$. We observe that in general, the corrected estimators perform much better than the plug-in. They have a smaller regret across all $n$, and for some $n$ there is even a factor of $5$ improvement. In addition, they converge faster to a regret close to $0$. Regarding the efficient $\hntcbd$, we see that despite the necessary relaxations, it has performance that is on par with both $\hntco$ and $\hntcbo$.

Next,  in Fig.~\ref{fig:conLevels} we plot the regrets averaged for the different dimensionalities of the joint probability distributions, i.e., $r_n(\tau, \cP^{2}_{[0.1,0.5]})$ (left), $r_n(\tau, \cP^{3}_{[0.1,0.5]})$ (middle), and $r_n(\tau, \cP^{4}_{[0.1,0.5]})$ (right). Under this different view, we  see that the plug-in estimator $\hntc$ has an increasing difficulty to converge to $0$ regret with respect to dimensionality, while the corrected estimators do not exhibit this behavior, as expected.  Among the corrected, the differences are more profound for $d=2$ with $\hntcbd$ having  worse performance. This ``artifact" can be attributed to the following behavior. For small $n$, not all $5$ random variables ($2$ dependent, $3$ independent) get to have samples with domain size $3$, and hence, $\hntcbd$ that penalizes with the product of domain sizes misses the $2$ dependent variables when they are sampled with domain size $3$, but the independent ones with domain size $2$. In addition, for $d=2$ the maximum is obtained for the pair of the dependent variables, with its subsets having a score of $0$ (since they are singletons). We do not observe this behavior for $d=3,4,$ for the simple fact that the subsets have a non-zero score, hence contributing to better regret. 

Finally, in Fig.~\ref{fig:conStrengths} we plot the regrets averaged over two ``strengths" of correlation, low with $p^{(i)}\in r_n(\tau, \cP^{[2,4]}_{[0.1,0.3)})$ (left) and relatively high $p^{(i)}\in r_n(\tau, \cP^{[2,4]}_{[0.3,0.5]})$ (right). Again, the corrected estimators have better regret curves. Since their correction is based on a null hypothesis model, they are particularly well-suited for the scenario where the correlation is low, i.e., closer to independence. The plug-in $\hntc$ on the other hand, cannot distinguish between the chance effects, and hence, has an almost flat curve as we can see in the left plot. However, even where there is better separation with such effects, the corrected estimators still outperform the plug-in. 

Overall, we see that our proposed corrected-for-chance estimators $\hntco, \hntcbo,$ and $\hntcbd,$ clearly outperform the plug-in, sometimes even by a factor of $5$. In addition, we observe that the efficiently computable $\hntcbd$  has statistical properties that are on par with $\hntco$ and $\hntcbo$.  

\subsection{Optimization performance}\label{sec:algoper}

In this section we investigate the performance of the bounding function $\btcref$ and algorithms proposed for exhaustive (\textbf{$\bnb$}) and heuristic search (\textbf{$\BEAM$}) for the reliable normalized total correlation $\hntcbd$. For the evaluation, we consider benchmark data from the KEEL data repository~\cite{keel}, and particularly all classification  datasets with no missing values and $d \geq 7$, resulting in $49$ datasets with $n \in [101,1025010]$ and $d \in [7,91]$, summarized in Table~\ref{tab:opt}. All metric attributes are discretized in 5 equal-frequency bins. This experiment is executed on a Intel Xeon E5-2643 v3 with 256 GB memory. Our code is online for research purposes.\!\footnote{\giturl}

We employ the  two algorithms in order to retrieve the top correlated set. For $\bnb$, we set $\alpha$ to be the highest possible in increments of $0.05$ such that it terminates in less than $30$ minutes, and report in Table~\ref{tab:opt} the runtime, the percentage of the pruned search space,\!\footnote{defined as $100-(100*q)/2^d$, where $q$ are the nodes $\bnb$ explored} the depth of the solution, the maximum depth $\bnb$ had to selectively reach, and the quality $\hntcbd$ of the top correlated set. For $\BEAM$ we report runtime and the difference of the quality for the top result with that from $\bnb$. We average runtimes over $3$ independent executions. 

We observe that $\bnb$ is highly efficient as it finds the optimum solution in less than $30$ minutes (i.e., $\alpha=1$) for $42$ out of $49$ datasets. In $30$ of them, it takes less than a minute. For all $49$, it requires $77$ seconds on average. The bounding function $\btcref$ is very effective in pruning, enabling the discovery of optimum solutions on datasets such as \textit{coil2000} and \textit{move. libras} with $86$ and $91$ attributes, that with exhaustive search would otherwise be impossible. In addition, an average of $5$ maximum depth combined with an average solution size of $2.2$, shows that the synergy of $\btcref$ and enumerated search space allows to selectively explore based on the structure of the data, and not simply by cardinality. That is, it can potentially go to higher levels for promising candidates. 

The $\BEAM$ algorithm requires only a couple of seconds on the majority of the datasets. On average, it terminates after $3$ seconds. In addition, the solutions produced by $\BEAM$ are almost optimal considering that there are only $2$ negligible cases where the two algorithms differ. In general, for a solution on the second level $\BEAM$ cannot ``stray" enough. We do observe, however, that even for solution cardinalities of $3$ and $4$, $\Beam$ solutions are identical to those of $\bnb$.

Overall, both algorithms are very effective with $\btcref$ as a bounding function. The $\bnb$ algorithm would be preferable in scenarios were solution guarantees are required, while $\BEAM$ when efficiency is more important, e.g., on very large datasets.

\subsection{Example discoveries}\label{sec:exam}

Last, we proceed with presenting concrete correlated sets discovered on two applications: finding correlations associated with win/loss on Tic-tac-toe, and identifying sets of co-inhabitant European land mammals together with factors affecting their coherence.

\textbf{Tic-tac-toe} is a game of two players where each player picks a symbol from $\lbrace x,o \rbrace$ and, taking turns, marks his symbol in an unoccupied cell of a $3\times3$ game board. A player wins the game if he marks 3 consecutive cells in a row, column, or diagonal. A game can end in draw if the board configuration does not allow for any winning move. The dataset consists of 958 end game, winning configurations, i.e., there are no draws. There are $10$ input variables $\cI=\{X_1, \dots, X_{10}\}$, where $X_i,i \in [1,9]$ represent the cells of the board, taking values in $\lbrace{ x,o,b \rbrace}$ with $b$ denoting an empty cell, and $X_{10}$ is the binary outcome of the game for player with symbol $x$.

We present in Fig.~\ref{fig:tic} the top-$9$ results retrieved with $\hntcbd$. The input variables $X_i, i\in[1,9]$ are mapped to their corresponding board positions and color indicates the result. Red designates the result set contains $X_{10}$. We observe that top-$1,2,8,9$ are all winning configurations, and top-$3$ has $X_5$ from which the majority of winning configurations go through. Top-$4,5,6,7$ are losing configurations, something that can be validated by superimposing, for example, top-$1$ and top-$4$. The blue results also appear to be four rotations of a unique configuration, indicative of a potential common losing pattern. In a nutshell, $\hntcbd$ identifies interesting ``red" and ``blue" correlated sets that can act as latent factors for win and loss, respectively. 

Regarding $X_{10}$, we should be expecting correlation with the losing configurations in a similar manner as the winning ones. This can be attributed to the fact that the losing configurations are in general more ``random" compared to winning, and this combined with the small size of the dataset, cannot support a ``losing" top result of size $4$.

As a further experiment, we use estimators $\hntc,\hntco,\hntcbo$  with exhaustive search. We report that $\hntc$ essentially orders the results according to cardinality, i.e., the top-$1$ is all the input variables $\cI$, the next $9$ are all subsets of $\cI$ with size $9$ etc. For $\hntco$ and $\hntcbo$ there is agreement with the top $4$ of $\hntcbd$, but the next $5$ are all supersets of the top $2$ with an extra cell. We find the results of $\hntcbd$ to be more interesting in this case.

Lastly, we note that the nature of this game implies that the cells are roughly independent, i.e., $p(X_1, \dots, X_9) \approx \prod_{1}^{9}p(X_i)$, and that subsets of these cells should become dependent the moment they are conditioned on $X_{10}$. However, they can take any of $3$ values and hence, any dependence is expected to be small. For example, the top-$1$ of $\hntcbd$ has score $0.08$, and when measured with the plug-in $\hntc$, has a score of $0.12$. These two values are more indicative for the maximum amount of correlation we should expect, in contrast to the value $0.36$ for the top-$1$ retrieved with $\hntc$. To put it differently, $\hntcbd$ is able to identify aspects of the ``low" signal residing in this dataset.

We now shift our attention into data that contain a lot more information, and particular the \textbf{European land mammal} dataset~\cite{heikinheimo2007mammals}. The dataset contains presence/absence records of $124$ land mammals for a set of $2183$ grid cells covering Europe, where each cell is approximately $50\times50$ km. The dataset also contains enviromental information, such as temperature, precipitation, and elevation, which we discretize into $2$ categories to reflect low and high.

In the top results we mainly recover coherent sets of mammals that are categorized as small, i.e., in the families of Insectivora, Rodentia, and Lagomorpha, and are endemic in southern Europe and the European Alps. For example, the top-$1$ set with score $0.7$ contains the Cretan spiny mouse and the Cretan shrew, and top-$2$ with same score the Savi's pine vole and Crested porcupine, both rodents inhabiting Italy. Larger sets include various species of shrews and rodents. Particularly interesting is the set of the greater white-toothed shrew, the Canarian shrew, and the Osorio shrew. The latter two appear mainly in the Canary islands, while the former in central-west Europe. This set could be used, for example, as an indicator that Osorio shrew, originally described as a separate species, indeed belongs to the shrew family~\cite{molina2003shrew}. Furthermore, we find that the coherence of sets with large mammals depends on the presence of environmental information.  As an example, a set with score $0.45$ contains two large mammals, moose and Arctic fox, along with three rodents, wood lemming, Norway lemming, and grey red-backed vole. All these inhabit Scandinavia. More coherent sets of large mammals appear together with environmental information, e.g., the set temperature, moose, European bison, and wild goat, with score $0.37$. We find that our analysis is to a large extend in sync with that of Heikinheimo et al., and particular the coherent sets of small mammals in southern/central Europe, and the environmental effect on the coherence of sets with large mammals~\cite{heikinheimo2007mammals}.

	\section{Conclusion}\label{sec:conc}
We considered the problem of measuring and efficiently discovering interpretable correlated sets from data. We adopted an information theoretic approach, and proposed a reliable and efficient estimator for normalized total correlation. In addition, we proposed effective algorithms for exhaustive and heuristic search, enabled by a tight bounding function.

Regarding future work, we see many possibilities for extensions and improvements. First, a similar framework could be derived by finding other suitable estimators, e.g., based on parametric solutions~\cite{vinh2014reconsidering}, and then developing efficient algorithms for these estimators.  Second, using a conditional version of normalized total correlation would allow the discovery of correlated sets with respect to control variables, e.g., for fairness. As an application, we could control with the top results discovered in subsequent executions of the algorithm and retrieve increasingly diverse results.

Regarding the algorithmic part, it could be possible to extend the NP-Hardness proof of Mandros et al.~\cite{mandros2018discovering}, and show that the optimization problem under consideration is also NP-Hard. Moreover, the recent algorithmic framework of Pennerath~\cite{pennerath2018efficient} for computing entropic measures, could potentially be applied here to efficiently discover results for larger $k$ values.

	
	\bibliographystyle{IEEEtran}
\begin{table*}[htbp]
  \centering
  \caption{\textbf{Datasets used in Sec.\ref{sec:algoper} together with the results of the experiment}. The $\alpha$ values correspond to the maximum possible approximation guarantee in increments of $0.05$ such that branch-and-bound ($\bnb$) finishes in less than $30$ minutes. Maximum search level is the maximum level that $\bnb$ had to selectively reach in order to find the solution, while solution depth is the depth where the solution was found. Pruning percentage is the amount of search space reduced by the bounding function and $\bnb$. The last two columns correspond to the value of the top solution retrieved by $\bnb$, and the difference with the value of the top solution by $\BEAM$, respectively.}
        \begin{tabular}{lrrlccrrrcc}

        	& & & & \multicolumn{2}{c}{search level} & & 	\multicolumn{2}{c}{time(s)} & \multicolumn{2}{c}{$\hntcbd(\cX^*)$\hphantom{aad}}  \\         	\toprule
        	dataset & \#rows  & \#attr. & $\alpha$ & max  & sol. & prune\% & \bnb & \Beam & \bnb & $\bnb-\Beam$ \\ \midrule
    abalone & 4174  & 9     & 1     & 6     & 2     & 48.90  & 0.5   & 0.2   & 0.67  & 0 \\
    appendic. & 106   & 8     & 1     & 3     & 2     & 71.37 & 0.1   & 0.1   & 0.56  & 0 \\
    australian & 690   & 15    & 1     & 3     & 2     & 99.67 & 0.1   & 0.1   & 0.97  & 0 \\
    bupa  & 345   & 7     & 1     & 5     & 2     & 15.70  & 0.1   & 0.1   & 0.10   & 0 \\
    car   & 1728  & 7     & 1     & 5     & 2     & 14.87 & 0.1   & 0.1   & 0.20   & 0 \\
    chess & 3196  & 37    & 1     & 9     & 3     & 99.99 & 617.4 & 0.6   & 0.64  & 0 \\
    coil2000 & 9822  & 86    & 1     & 3     & 2     & 99.99   & 7.2   & 6.7   & 0.99  & 0 \\
    connect & 67557 & 43    & 0.8   & 6     & 2     & 99.99 & 1094.8 & 11.5  & 0.62  & 0 \\
    contracept. & 1473  & 10    & 1     & 6     & 2     & 50.59 & 0.3   & 0.1   & 0.25  & 0 \\
    fars  & 100968 & 30    & 1     & 2     & 2     & 99.99 & 15.4  & 10.3  & 0.99  & 0 \\
    flare & 1066  & 12    & 1     & 4     & 2     & 93.36 & 0.1   & 0.1   & 0.62  & 0 \\
    german & 1000  & 21    & 1     & 6     & 2     & 98.63 & 15.8  & 0.1   & 0.26  & 0 \\
    glass & 214   & 10    & 1     & 5     & 2     & 58.57 & 0.1   & 0.1   & 0.19  & 0 \\
    heart & 270   & 14    & 1     & 5     & 2     & 83.33 & 0.4   & 0.1   & 0.17  & 0 \\
    ionosphere & 351   & 34    & 1     & 5     & 2     & 99.99 & 69.8  & 0.1   & 0.45  & 0 \\
    kddcup & 494020 & 42    & 1     & 4     & 2     & 99.99 & 284.4 & 73.5 & 0.98  & 0 \\
    kr-vs-k    & 28056 & 7     & 1     & 5     & 3     & 8.26  & 1.6   & 0.3   & 0.18  & 0 \\
    led7digit & 500   & 8     & 1     & 6     & 2     & 37.50  & 0.1   & 0.1   & 0.50   & 0 \\
    letter & 20000 & 17    & 1     & 8     & 2     & 80.37 & 390.2 & 1.2   & 0.41  & 0 \\
    lymph. & 148   & 19    & 1     & 6     & 2     & 99.15 & 0.5   & 0.1   & 0.28  & 0 \\
    magic & 19029 & 11    & 1     & 5     & 2     & 81.63 & 2.5   & 0.3   & 0.67  & 0 \\
    monk  & 432   & 7     & 1     & 4     & 2     & 32.23 & 0.1   & 0.1   & 0.31  & 0 \\
    move. libras & 360   & 91    & 1     & 3     & 2     & 99.99   & 12.7  & 0.5   & 0.92  & 0 \\
    nursery & 12690 & 9     & 1     & 4     & 2     & 68.19 & 0.6   & 0.2   & 0.60   & 0 \\
    optdigits & 5620  & 65    & 0.35  & 2     & 2     & 99.99   & 3.3   & 3.4   & 0.49  & 0 \\
    page  & 5472  & 11    & 1     & 5     & 2     & 77.71 & 0.8   & 0.1   & 0.69  & 0 \\
    penbased & 10992 & 17    & 1     & 7     & 3     & 85.38 & 118   & 0.8   & 0.51  & 0 \\
    poker & 1025010 & 11    & 0.9   & 8     & 4     & 4.95  & 1760.8 & 20.6  & 0.02  & 0 \\
    ring  & 7400  & 21    & 0.1   & 4     & 2     & 99.93 & 4.4   & 0.4   & 0.08  & 0 \\
    saheart & 462   & 10    & 1     & 5     & 2     & 52.95 & 0.1   & 0.1   & 0.21  & 0 \\
    satimage & 6435  & 37    & 0.65  & 6     & 4     & 99.99 & 632.8 & 1.6   & 0.55  & 0.004 \\
    segment & 2310  & 20    & 1     & 5     & 2     & 99.71 & 2.4   & 0.1   & 0.82  & 0 \\
    shuttle & 58000 & 10    & 1     & 7     & 4     & 57.00    & 16.2  & 1.4   & 0.58  & 0 \\
    sonar & 208   & 61    & 1     & 5     & 2     & 99.99 & 1246  & 0.2   & 0.35  & 0 \\
    spambase & 4597  & 58    & 1     & 4     & 2     & 99.99 & 130.6 & 2.0     & 0.89  & 0 \\
    spectf. & 267   & 45    & 1     & 5     & 2     & 99.99 & 331.9 & 0.1   & 0.29  & 0 \\
    splice & 3190  & 61    & 0.25  & 2     & 2     & 99.99 & 1.4   & 1.5   & 0.25  & 0 \\
    texture & 5500  & 41    & 1     & 3     & 2     & 99.99 & 1.4   & 1.4   & 0.99  & 0 \\
    thyroid & 7200  & 22    & 1     & 6     & 2     & 99.67 & 26.5  & 0.5   & 0.40   & 0 \\
    tic-tac-toe   & 958   & 10    & 1     & 7     & 4     & 11.04 & 0.4   & 0.1   & 0.08  & 0.005 \\
    twonorm & 7400  & 21    & 0.2   & 6     & 2     & 99.13 & 84.1  & 0.4   & 0.13  & 0 \\
    vehicle & 846   & 19    & 1     & 4     & 2     & 99.79 & 0.4   & 0.1   & 0.87  & 0 \\
    vowel & 990   & 14    & 1     & 2     & 2     & 99.43 & 0.1   & 0.1   & 0.95  & 0 \\
    wdbc  & 569   & 31    & 1     & 4     & 2     & 99.99 & 0.9   & 0.2   & 0.90   & 0 \\
    wine  & 178   & 14    & 1     & 4     & 2     & 93.19 & 0.1   & 0.1   & 0.48  & 0 \\
    wine-red & 1599  & 12    & 1     & 6     & 2     & 53.13 & 2.1   & 0.1   & 0.25  & 0 \\
    wine-white & 4898  & 12    & 1     & 7     & 3     & 51.29 & 6.0     & 0.3   & 0.32  & 0 \\
    yeast & 1484  & 9     & 1     & 5     & 2     & 64.21 & 0.1   & 0.1   & 0.19  & 0 \\
    zoo   & 101   & 17    & 1     & 4     & 2     & 99.87 & 0.1   & 0.1   & 0.79  & 0 \\ \bottomrule
    avg.   & 39000 & 25    & 0.92  & 5   & 2.2   & 77.00 & 142   & 3     &       &  \\
    \end{tabular}%
  \label{tab:opt}%
\end{table*}%

\end{document}